# Predicting Country Instability Using Bayesian Deep Learning and Random Forest

Adam Zebrowski and Haithem Afli
ADAPT Centre
*Munster Technological University*
Cork, Ireland
*Name.FamilyName@adaptcentre.ie*

*Abstract*— **Country instability is a global issue, with unpredictably high levels of instability thwarting socio-economic growth and possibly causing a slew of negative consequences. As a result, uncertainty prediction models for a country are becoming increasingly important in the real world, and they are expanding to provide more input from 'big data' collections, as well as the interconnectedness of global economies and social networks. This has culminated in massive volumes of qualitative data from outlets like television, print, digital, and social media, necessitating the use of artificial intelligence (AI) tools like machine learning to make sense of it all and promote predictive precision [1]. The Global Database of Activities, Voice, and Tone (GDELT Project) records broadcast, print, and web news in over 100 languages every second of every day, identifying the people, locations, organisations, counts, themes, outlets, and events that propel our global community and offering a free open platform for computation on the entire world. The main goal of our research is to investigate how, when our data grows more voluminous and fine-grained, we can conduct a more complex methodological analysis of political conflict. The GDELT dataset, which was released in 2012, is the first and potentially the most technologically sophisticated publicly accessible dataset on political conflict.**

## I. INTRODUCTION

In social science research, social and political events detection is a well-known, critical and difficult activity. Detecting the occupy demonstration case (OPE), which typically campaigns against social, political, and economic injustice, is of special importance. People protest against problems that affect their life and for which they believe the government (local, state, or national) is responsible during the demonstration (e.g., unfavourable election, poor infrastructure, etc.). Identifying the participants' contact patterns and estimating the likelihood of an OPE will serve as a guide for government emergency response. In recent years, computational scientists have had access to a plethora of data tools [9]. Among them, the Global Dataset of Incidents, Place, and Tone (GDELT), built on open-source data, comprised over 300 million machine-coded events in near real-time (e.g. every day, every fifteen minutes). This dataset has been used to find trends in a variety of incidents, including domestic political conflicts, natural disasters, racial and religious conflict, and more. In a graph-based approach, a monthly time window was used to forecast domestic political crises, which is a much too large resolution for real-time information systems.

We used the GDELT dataset, which includes machine-encoded archives of international events originating from news reports, to collect bilateral sequences of inter-country events and a Bayesian standard mining algorithm to find norms that best represented the observed behaviour. A statistical study found that a probabilistic model with explicit normative reasoning outperformed a reference probabilistic model in terms of data matching. The Global Archive of Events, Language, and Tone (GDELT) is a continuously updated geopolitical activity database of over half a billion records. The most recent version, GDELT 2.0, is free and open source, and it is updated every 15 minutes. The database contains an incidents table of 60 attributes for each incident (such as the type of the accident and the countries involved), and it has been used for research such as estimating future levels of unrest in Afghanistan, Pakistan, Thailand among others, and identifying protest activities around the world [10].

In this work we aim to better understand the current state of political uncertainty, considering the dynamics of a quickly evolving and increasingly intertwined global climate on nation instability levels [11]. This would be supported by the use of Bayesian inference to provide an in-depth learning approach capable of dealing with such a dynamic and changing stability environment. The massive data sources from GDELT, the World Bank, and the Correlates of War on individuals, locations, incidents, and actions can be combined with A.I. research capability to promote a systemic view of the global stability that considers the different networks and aspects that can affect insecurity [12].

Our main objectives in this study are the following:

- To look at countries that have a lot of contextual data that has been gathered over a long period of time. Wide data density, precise Geo-coding, Sub-state, latitudinal, and longitudinal data are all supposed to be crucial background specifics.
- To assess the benefits and shortcomings of GDELT machine-coding in order to ensure the precision and credibility of the final predictive model [13].
- To create a model that uses deep learning, Bayesian techniques, and random forest techniques to enable successful intra-county and location-specific analyses, as well as predictive validity in the context of country stability.
- Examine cutting-edge implementations of deep learning approaches to big data in forecasting, as well as best practices and novel predictive analytics applications and processes [14].

## II. RELATED WORK

### A. Uncertainty and Complexity in AI Models)

Artificial Intelligence (AI) models were found to be quite capable of accurate predictive analysis and obtaining an early indication of country instability from limited information [2, 3]. Such models have become increasingly important considering that classical predictive models have found it increasingly difficult to forecast instability due to evolving mechanisms and factors influencing country instability, such as increased global interconnections of economies, trade and communications [4, 5, 6]. It is important to discuss the concept of uncertainty and complexity management in AI systems and their importance in developing more sustainable and effective prediction systems. The GDELT datasets to be used in this research, which will be complemented by World Bank, Correlates of War, etc., have been categorised using AI techniques. A review of the underlying AI algorithms employed in the categorisations will be undertaken as well as examining the associated ethical and human-machine issues pertaining to machine learning and AI techniques and algorithms.

Prediction systems can significantly benefit from increased capability to handle uncertainty and complexity, since these qualities are inherent and widespread in real-world settings. The ability to cope with these is therefore critical in developing superior intelligence in machine learning applications [7]. Different from risk, which can be evaluated statistically, uncertainty is characterised by information that may be incomplete, random, ambiguous or inconsistent [7, 8]. Complexity, on the other hand, involves the complicated or intricate interaction of different parts of a system. While the capability for increased computational or processing power may be helpful in decision-making involving complexity, uncertainty requires some level of generalisation, which provides the ability to be able to understand and potentially cope with unknown situations and new data [9]. Deep learning algorithms have gained a significant following in the machine learning field for their superior performance concerning generalisation while also having the capacity to handle complex abstractions in analysing vast amounts of diverse big data [10, 11].

### *B. Bayesian Deep Learning*

Connecting Bayesian probability theory with deep learning has been shown to improve its efficiency and minimise challenges of overfitting and computational efficiency. Dropout techniques, which randomly reduce certain DNN connection weights to zero to improve computing performance and reduce overfitting, have been shown to be identical to approximate inference in Bayesian Modelling. The inherent capabilities of Bayesian probability theory also help to improve the capability of DNNs to handle uncertainty and learning from small data domains [18, 19]. The introduction of Bayesian inference has been shown in studies to speed up the process of adjusting the model, dimensions and weights in the adaptation of more sparse or fit-for-purpose deep neural networks, requiring less computation and increasing performance. An empirical study of an adaptive Bayesian (AEB) sparse deep learning method adapting MNSIT with shallow convolutional neural networks (CNNs) increased its compression performance and improved the system's resistance to adversarial attacks [26]. Another study based on Parsimonious deep Neural Networks that combined Bayesian non-parametrics with a forward model selection strategy provided adaptive hidden layers whose number of active units are automatically adjusted based on actual data, therefore reducing excessive dimensionality which consumes computation time [19].

## III. Country Instability Prediction Based on GDELT

Existing literature shows that the application of machine learning, Bayesian inference and deep neural networks have been applied to GDELT and related data for predicting country instability with varying results. Working with a Bayesian theoretical framework and a big open-source dataset, a fused-model approach was used to increase the ability to improve prediction precision for civil unrest — especially with information from heterogenous sources, such as using datasets from multiple countries when compared to models using information from only an individual country [27]. A convolutional neural network method was applied to moderate-resolution imagery in Nigeria to generate measures of poverty and development in developing countries. Findings from this study indicate that CNN-based methods could be used to estimate low-cost development related information which, however, does not replace traditional analysis [28]. On the other hand, a comparative study of machine learning techniques showed the random forest algorithm outperforming the Naive Bayes algorithm. In summary, there appears to be very few studies available utilising the recently increased capabilities of deep learning networks to provide high performance predictions due to increased availability of big data and computational capacity [12, 13]. This provides a gap that we explore in this paper.

### *A. Limitations of GDELT Data*

*It will be important when working with GDELT and related datasets to understand the limitations that are likely to influence results and which need to be addressed in the design of the research methodology. One of these is the fact that GDELT's categorised information presents a black box due to the use of non-transparent algorithms for organising available information. As a result, it is not very clear how many media outlets are monitored in the datasets [20]. The theme classifications provided can also be difficult to contextualise. As a result, the UK office of National Statistics has declared, for example, that this makes the identification of UK-based disasters from GDELT unreliable as well as numeric information reporting deaths associated with the disaster of interest. It is therefore important that activities are undertaken to augment the quality of the information obtained from GDELT. In addition to the more diverse and nuanced data introduced by the use of World Bank and War Correlations, external validity tests using parallel data will be important. This could also include the use of maps and satellite images, as used in the empirical model construction of future conflict in Afghanistan [21].*

### *B. Ethics and Human Computer Interaction AI Issues*

Avoid Considering the relatively vague mechanism presented by deep neural networks built primarily on heuristics and the relatively immature stage of development in the area, it is important to consider how this impacts the research from an ethical and legal perspectives. This also applies to the use of GDELT datasets, which are primarily proprietary and non-transparent [22]. Steps need to be taken when constructing algorithms to reduce potential legal exposure due to hidden biases embedded either in datasets used or within the elements of the AI model. This includes ensuring diversity in the provision of training data — especially with the proposed use of information from a hybrid of European countries to avoid unexpected false results and in-built discrimination [23, 24].

Human-computer interaction issues are also of interest, as it is not yet clear why deep learning networks work so well, since much of their development over decades have been based on mimicking nature and using experience and trial and error to improve the system. One of the most important aspects of human-to-human interactions is the explainability of actions and at least an understanding of underlying reasons. These are important for creating sustainable intelligent systems which can provide high confidence of reliability and protection from ethical and legal issues [25]. It is thus important that these issues are addressed to avoid over-reliance on inexplicable internal workings of an AI system which may lead to an unexpected and significant black swan failure due to its black box mechanism.

### C. *GDELT event database*

The GDELT Event Database tracks over 300 types of physical activity around the world from demonstrations and protests through peace appeals and international exchanges. As a result, the current archive contains more than 2.5 terabytes of data per year. In terms of absolute figures, it has over a quarter billion data [16]. The website, which is powered by Google Jigsaw, contains data from 1979 to the present and is updated every 15 minutes as of April 1, 2013. In other words, documents are constantly being added to the database. The enormous number of event logs - more than any other event dataset - provides a new insight on this field of study. So far, few studies have attempted to use GDELT to forecast civil conflict, and only a few researchers have used GDELT to make forecasts.

The Global Database of Incidents, Language, and Tone (GDELT) is a modern CAMEO-coded dataset that contains geolocated events from 1979 to the present with global scope. The information was gathered from news accounts from around the world [15]. This dataset currently contains regular analysis of incidents contained in news stories released that day. The CAMEO taxonomy divides case forms into four categories: verbal cooperation and material cooperation (numbers 1 to 10) and verbal conflict and material conflict (numbers 11 to 20). Furthermore, each case has 32 separate responsibilities for the players, such as police forces, government, and military. Each record in GDELT contains details about a single incident. We use the following attributes from a case to build our models: "MonthYear, Actor1Type, Actor2Type, RootEventCode, AvgTone, and GoldsteinScale, where Actor1Type and Actor2Type store the role of the actors participating in the event", RootEventCode $\in \{1, \ldots, 20\}$ identifies whether this incident is cooperative or contradictory, AvgTone is a subtle indicator of an event's magnitude that serves as a surrogate for its influence, and the GoldsteinScale captures the event's impact on a country's stability.

## IV. Experiments

### A. *Data pre-processing*

The first part of the work is filtering and pre-processing GDELT data so that it can be in feature engineering and labelling. The following process is applied to the raw GDELT data.

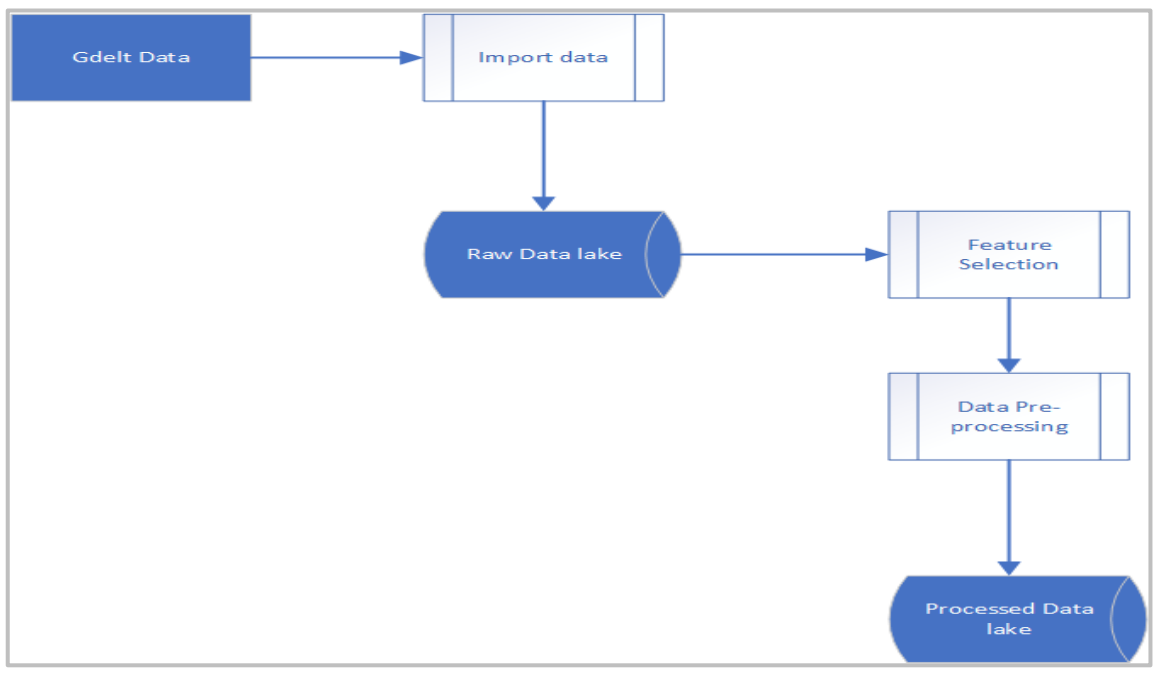

*Figure 1. Data pre-processing flow*

The data to be imported is stored in one master file that contains all pointers to 15-minute intervals payload files. Those files are zipped, hence during the import process those are unzipped as well as divided into yearly categorisation. At this stage all raw files are imported. Any subsequent import jobs run only on new data.

The output of this phase is set of files, one per country, that contain the relevant information to the analysis descried in this work.

### B. *Feature engineering*

After the data is pre-processed the next tasks of feature engineering and data labelling is performed. This process is described in the diagram below.

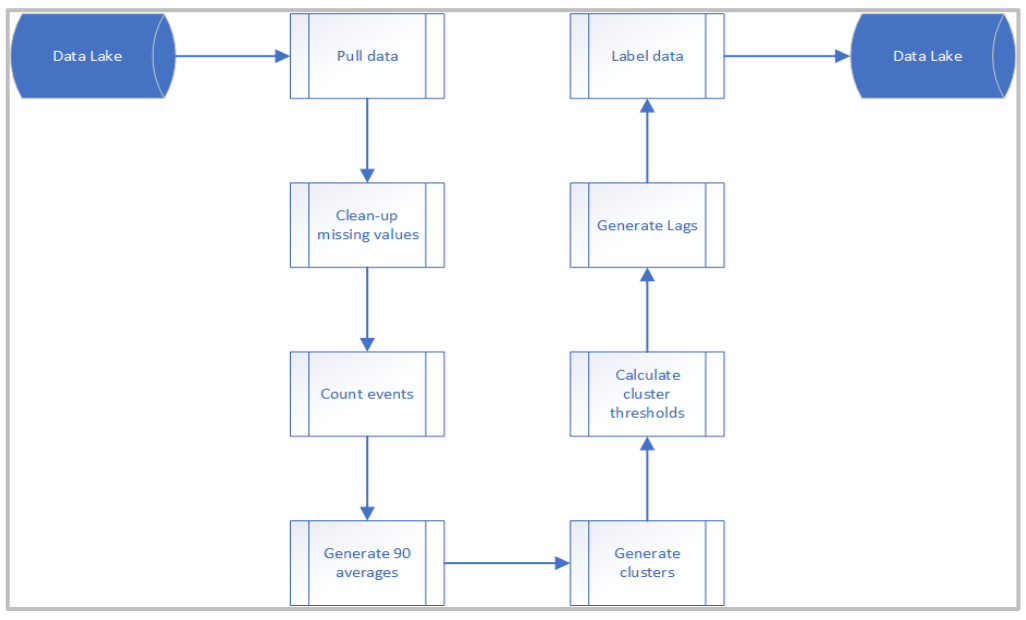

*Figure 2. Feature generation flow*

Data is pulled from the initial data lake and cleaned up for missing values. Given that GDELT is using machine algorithms to segregate the events, some of the data may be missing at this point. The next important step is to generate the count of events per day. GDELT creates one file every 15 minutes, hence there is a need to bring that data to daily granularity level.

Next important task is to acknowledge that social unrest does not happen on a single day and that process usually builds up gradually over an extended period of time. For this reason, the 90-day moving average is calculated as a base for clustering the events over time.

At this stage the mean clustering technique (MCT) is used to analyse the daily events. For the purpose of this work only 1 CAMEO code is used (#14), however that could be extended in the future work to not include root codes, but specific ones from 10, 11, 13 and 14 categorisations [16]–[17].

It is important to mention that the cluster calculations do not operate on the single days, but on the intervals. The in 3 to 7-day intervals were considered and the best results were obtained using 3-day intervals.

The key part of feature generation is to calculate the cluster thresholds for the particular events. The following formula is used to calculate:

$$\theta = MCTBarComp + 2.576 \times \sqrt{-((MCTBar - MCTBarComp)^2)}$$

This value is probably the most important one in feature generation as it will be used to segregate the events into possible unrest or not.

Where MCBarComp is calculated in the following manner:

$$MCTBarComp = -(MCT / MCTBar)$$

One the primary clustering is obtained; the next step is to add lags to the feature list to enable step ahead predictions. A lagging indicator is an apparent or detectable element that varies after the associated economic, financial, or market variable changes. Trends and shifts in trends are confirmed by lagging measures. Lag generation for rolling averages and future predictions is a naive and effective technique in time series forecasting. It can be used for data preparation, feature engineering, and even directly for making predictions. Hence, lags can be used to predict a variable in the sequence. Given that lags are interpolated from the existing data, they are prone to rapidly decreasing accuracy. For the purpose of this work the lag of maximum 7 days was used.

### C. *Labelling*

Once the preliminary processing is performed with respect to arrangement through the dates and clusters, the work can move to labelling the data. At this point it is wort mentioning that the analysis can be performed both on binary labelled data (focus of this work) as well as the threshold analysis (δ). Labels are generated based on cluster distance thresholds based on the following:

*MCT_Comp > δ -> unrest s assumed, otherwise it is not.*

Once this is executed, the processed and labelled data is then stored into a new CSV file for further analysis.

### D. *Classification*

The classification algorithm used here is a Random Forest algorithm, which is an ensemble learning-based supervised machine learning algorithm. The random forest algorithm incorporates many algorithms of the same kind, such as multiple decision trees, to create a forest of trees, thus the term "Random Forest." Both regression and classification tasks will benefit from the random forest algorithm. In the case of a regression problem, each tree in the forest predicts a value for Y for a new record (output). The final value can be determined by averaging all of the expected values from all of the trees in the forest. Alternatively, each tree in the forest predicts the group to which the new record belongs in the case of a classification query. Finally, the current record is given to the group that receives the greatest number of votes.

When you have both categorical and numerical elements, the random forest algorithm works well. When data has missing values or has not been scaled well, the random forest algorithm works well (although we have performed feature scaling in this article just for the purpose of demonstration). The code below extracts data for the countries mentioned below. We don't know which includes the data for which country right now. As a result, the first step removes both the code and the country name. The second step is focused on locating the region. The Random Forest algorithm must be trained so that it would be able to learn the patterns between the parameters and its values. Since the labelled data has been created, it will be used for training the classifier. The data is split into training dataset and testing dataset based on the dates. When Pakistan is considered, it contains 2222 instances of data out of which 65% of the data is split into the training data, while the remaining 35% is designated as testing data. The data split is based on the following principles:

- Events from 2015-2018 are used as training data;
- Events from 2019 are used as test data;
- Events from 2020 and 2021 are discarded to eliminate potential anomalies based on the outbreak of COVID-19 pandemic.

The trained data is sent to the classifier, where it learns the data and generates a training model. The random forest classifier now uses this training model to evaluate the algorithm and predict the stability of the testing dataset. The algorithm is evaluated with respect to accuracy and mean absolute error.

## V. Results

### A. *Performance Metric*

The learned model's efficiency is discussed in detail using an output metric. The intrusion's efficiency, behaviour, and actions in the network are evaluated using the Performance metric. According to the performance metric, the precision is 85 percent, with a mean absolute error of 0.15 degrees when looking at the current data and 75% when looking at interpolated data (looking ahead forecasting).

As a result, we can be certain that the algorithm has been sufficiently trained and that the final product is both accurate and effective.

### B. *Comparison of the Different Classifier Performance*

The proposed method produced excellent results, so we compared the accuracy of the random forest method to that of three other feature extraction classifiers to illustrate the applicability of the random forest method. The following methods were used as comparisons:

- Gaussian
- SVC
- KNN
- Decision tree
- Neural Network

As a result, we can conclude that the random forest approach is the best approach for this analysis. It not only produces stronger and more reliable results, but it also has a fast-training speed, which is critical in practice [26]. The below tables shows a detailed comparison of overall classifier results sorted by accuracy for 3 example countries: Pakistan, Egypt and Sudan:

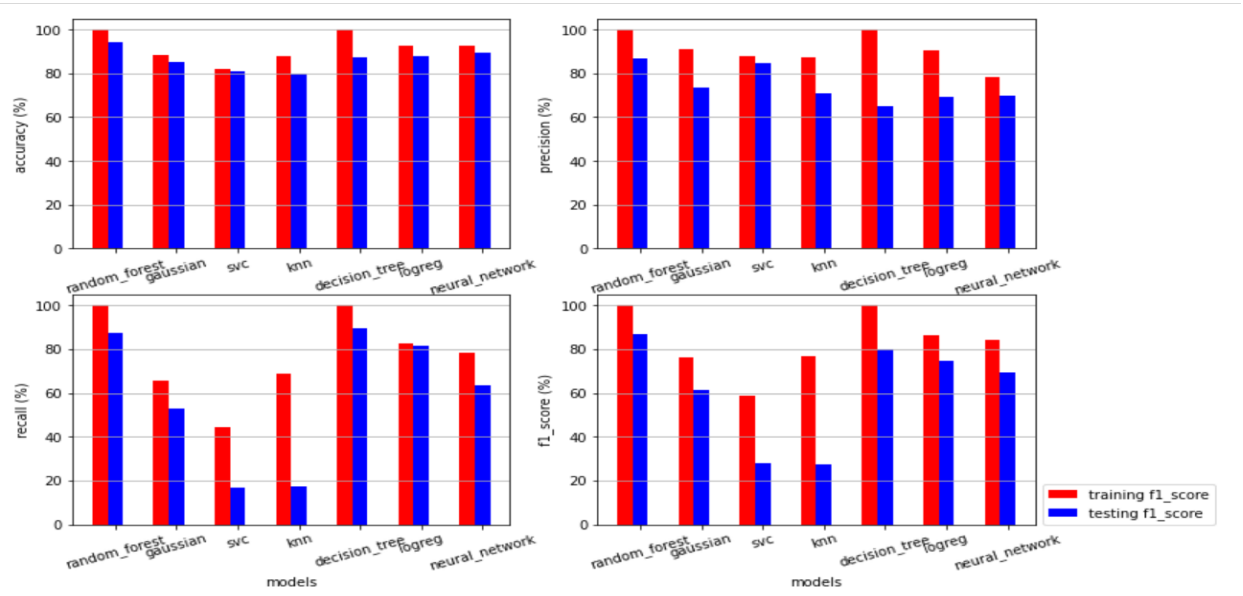

*Figure 3. Accuracy comparison for Pakistan*

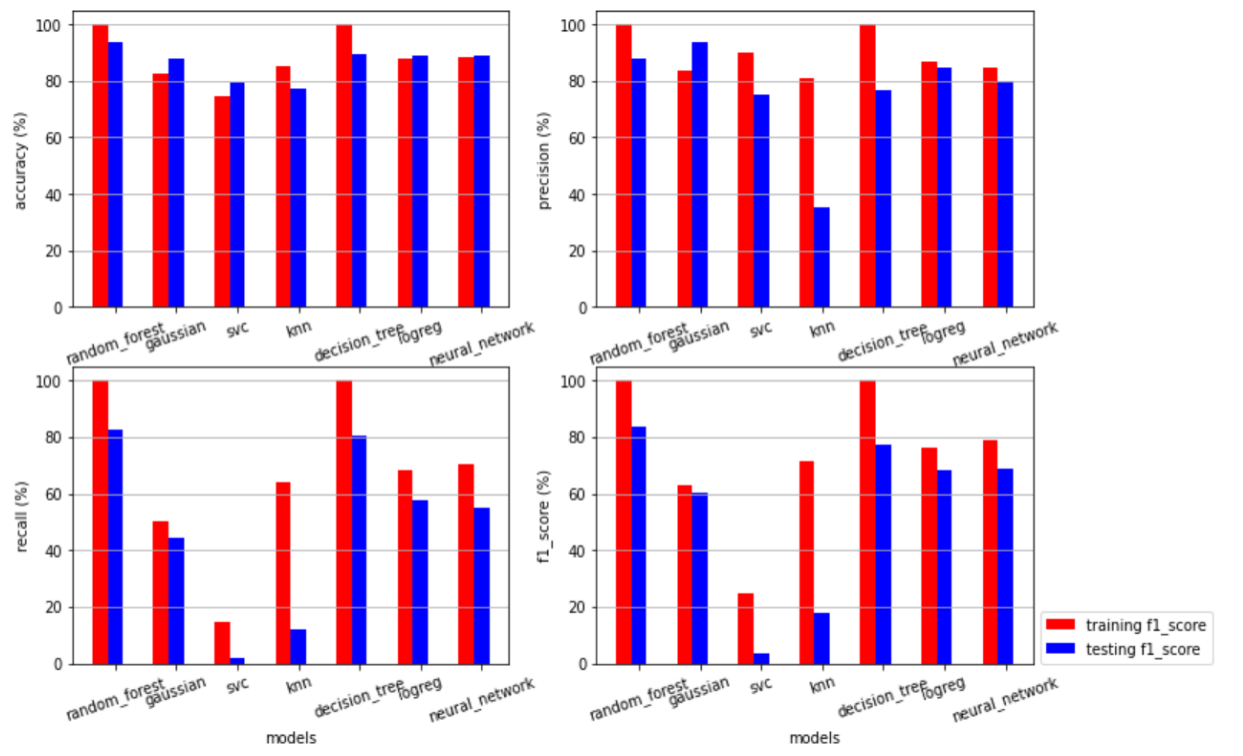

*Figure 4. Accuracy comparison for Egypt*

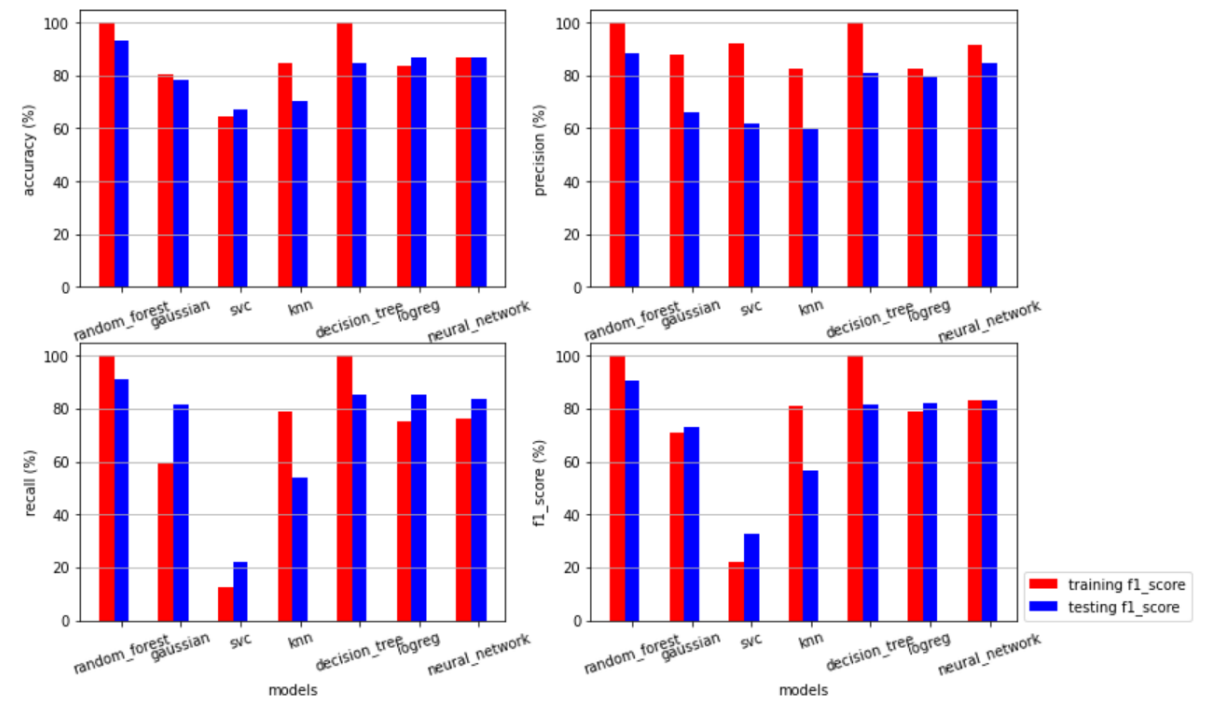

*Figure 5. Accuracy comparison for Sudan*

When compared to the other classifier models, the bar plots in Figures [9,10,11] clearly shows that my proposed model accuracy is very efficient and reliable. The rest of the models are lagging behind. As a result, we can conclude that the model can effectively define the probability of social unrest given the data and clustering applied.

## VI. Conclusion and Discussion

When The key contribution of the presented work is the creation of a framework for developing alternative economic and financial metrics that catch investor sentiments and topic popularity using GDELT, the Global Data on Events, Location, and Tone database, a free open portal that contains real-time world's radio, print, and web news. This research is being carried out as part of a project aimed at developing more accurate forecasting approaches for analysing the overall sector of a few countries. We have posted some early results from using this method to predict existing market conditions. This application shows that the technique works well at first, indicating that the method is right. Using the information obtained from the news media contained in GDELT, together with a deep Long Short-Term Memory Network opportunistically educated and tested using a rolling window system, we were able to achieve very good forecasting performance [27].

Policymakers could now use the experimental conflict case modelling technique applied to the GDELT dataset to map deteriorating or de-escalating circumstances in a country on a weekly or monthly basis. However, event-based frameworks will need further research to compensate for the databases' flaws, such as automatic data confirmation, new classifiers and dictionaries that represent the changing nature of violence, and, most importantly, evidence on the mechanisms between civil instability and violent conflict [28].

Working with GDELT requires a considerable amount of acquired informal knowledge about its quirks and shortcomings, which was mostly obtained from the large and involved GDELT user community. Future databases should be even more transparent about how their components are produced, especially for developments that go beyond standard actor and verb dictionaries. Although computer coding cannot absorb all of a specialist's expertise, it can be improved by scholars contributing to dictionaries and coding schema, resulting in potentially more useful datasets for researching political mobilisation. Datasets should preferably include URLs for all events and easily replicable coding schemes to enable end users to understand how the data was generated and the results of improvements in dictionaries, coding algorithms, and sources. To summarise, we conclude that while using vast volumes of machine-coded data to study phenomena such as civil society, political mobilisation, and government repression should be approached with care, it can be a useful analysis method. Indeed, the current widespread use of machine-coded event data for conflict forecasting can be expanded to include a much wider variety of comparative questions and approaches, which would be made possible by increased clarity and accuracy.

## Acknowledgment

This work is a part of the ITFLOWS project funded by the EU's Horizon 2020 research and innovation program under Grant Agreement No 882986 and the Science Foundation Ireland research Centre ADAPT through Grant 13/RC/2106_P2.